\documentclass[9.5pt,journal,compsoc]{IEEEtran}

\usepackage{amsfonts,amssymb}
\usepackage{amsthm}
\usepackage{array}
\usepackage{multirow}
\usepackage{color}
\usepackage{soul}
\usepackage{epsfig}
\usepackage{graphicx}
\usepackage{float}
\usepackage{mathtools}
\usepackage{tabularx} 
 \usepackage{booktabs}

\usepackage{lipsum}
\usepackage{setspace}
\usepackage{bm}
\usepackage{hyperref}
\usepackage{authblk}
\usepackage[final]{changes}
\usepackage{wrapfig}
\usepackage[cal=pxtx]{mathalfa}
\usepackage{rotating,threeparttable,booktabs,caption,dcolumn}
\usepackage{ragged2e}
\usepackage{enumerate}
\usepackage[linesnumbered,ruled]{algorithm2e}
\usepackage{dblfloatfix} 

\newcommand{\cmt}[1]{}

\newtheorem{theorem}{Theorem}

\newtheorem{assumption}{Assumption}

\newtheorem{lemma}{Lemma}
\newtheorem*{proof*}{Proof}

\newcounter{definition}[section]

\newcommand{\diver}{\mathrm{div}}

\author{Ashif Sikandar Iquebal$^\dagger$\thanks{$^\dagger$Corresponding author}}
\author{Satish Bukkapatnam}

\affil{Department of Industrial and Systems Engineering, Texas A\&M University}
\date{}

\ifCLASSOPTIONcompsoc
  \usepackage[nocompress]{cite}
\else
  \usepackage{cite}
\fi

\makeatletter
\def\footnoterule{\relax%
	\kern-5pt
	\hbox to \columnwidth{\vrule width 0.5\columnwidth height 0.4pt\hfill}
	\kern4.6pt}
\makeatother

\begin{document}
\title{Consistent estimation of the max-flow problem: Towards unsupervised image segmentation }

\IEEEtitleabstractindextext{%
	\begin{abstract}
		Advances in the image-based diagnostics of complex biological and manufacturing processes have brought unsupervised image segmentation to the forefront of enabling automated, on the fly decision making. However, most existing unsupervised segmentation approaches are either computationally complex or require manual parameter selection (e.g., flow capacities in max-flow/min-cut segmentation). In this work, we present a fully unsupervised segmentation approach using a continuous max-flow formulation over the image domain while optimally estimating the flow parameters from the image characteristics. More specifically, we show that the maximum \textit{a posteriori} estimate of the image labels can be formulated as a continuous max-flow problem given the flow capacities are known. The flow capacities are then iteratively obtained by employing a novel Markov random field prior over the image domain. We present theoretical results to establish the posterior consistency of the flow capacities. We compare the performance of our approach on two real-world case studies including brain tumor image segmentation and defect identification in additively manufactured components using electron microscopic images. Comparative results with several state-of-the-art supervised as well as unsupervised methods suggest that the present method performs statistically similar to the supervised methods, but results in more than 90\% improvement in the Dice score when compared to the state-of-the-art unsupervised methods.

	\end{abstract}

	\begin{IEEEkeywords}
		Continuous max-flow, unsupervised image segmentation, maximum {a posteriori} estimation, posterior consistency
\end{IEEEkeywords}}

\maketitle



%
\IEEEpeerreviewmaketitle

\IEEEraisesectionheading{\section{Introduction}\label{sec:introduction}}

\IEEEPARstart{T}HE objective of image segmentation is to partition an image into semantically interpretable and spatially coherent entities featuring similar characteristics, e.g., pixel intensities and texture \cite{reed1990segmentation}. Although, the history of image segmentation can be dated back to almost half a century ago, recent advances in the image-based diagnostics of complex biological processes \cite{wang2018deepigeos}, materials characterization \cite{duval2014image}, object tracking \cite{borji2012adaptive,boykov2001fast}, etc. have brought image segmentation to the forefront of enabling automated, on the fly decision making. 

\begin{figure}
	\centering
	\includegraphics[width=0.45\textwidth]{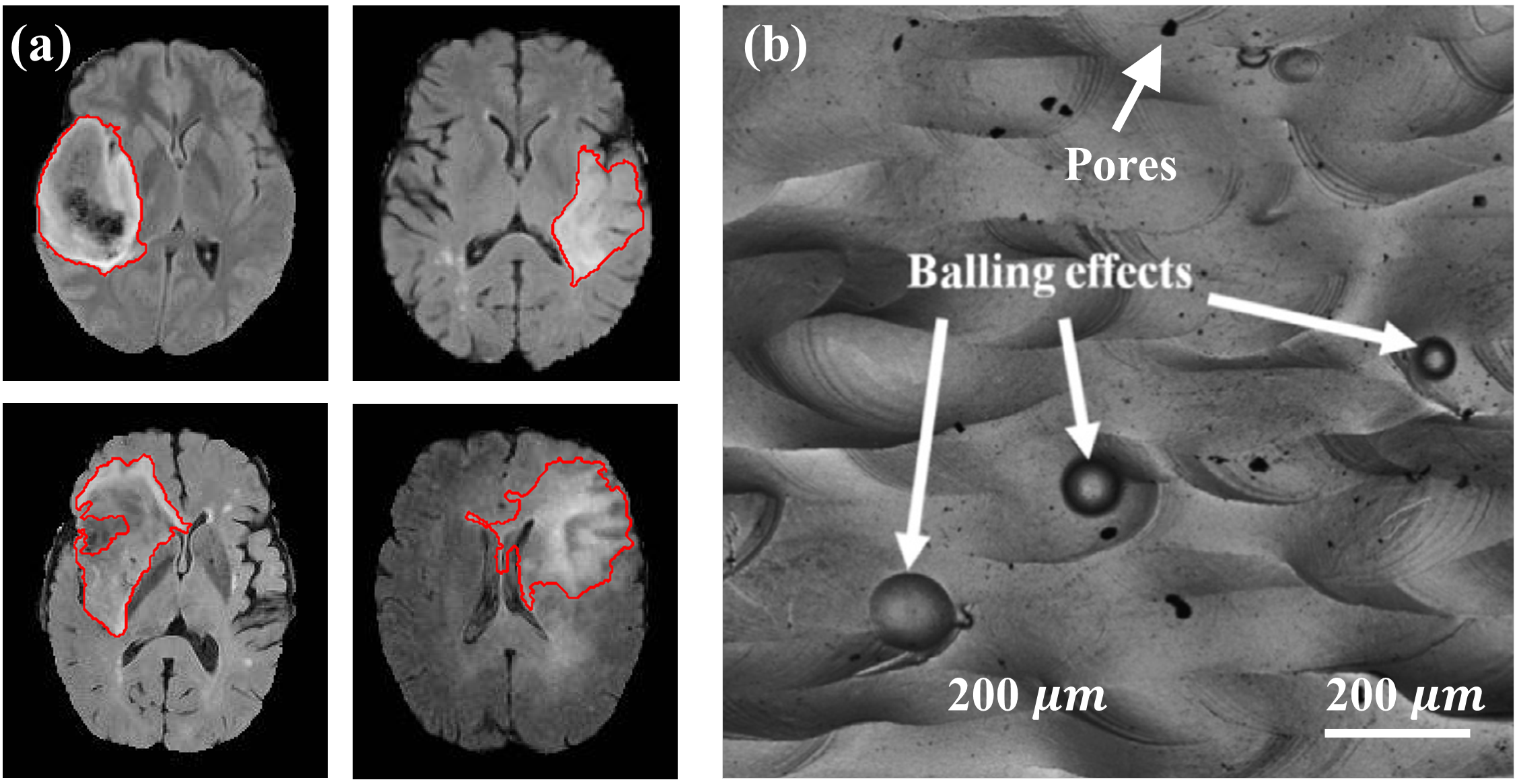}
	\caption{(a) MR scan of the brain slices showing the morphology of tumorous cells and (b) the morphology and concentration of defects (pores and balling effect--see Section 6.3 for more details) \cite{attar2014manufacture} in an AM component. These images reflect the uncertainty involved in the morphology, location and the pixel intensity distribution of the ROIs.}
	\label{fig:1}
\end{figure}

A vast majority of the current image segmentation approaches are supervised, i.e., they employ models that learn from previously labeled images to determine the partitions and/or the ROIs in a given image. Their performance relies on the availability of a large collection of labeled images, and at times, the expert knowledge \cite{koch2018multi}. With the growing database of images from newer imaging technologies coupled with the increasing emphasis on on the fly detection of novel (i.e., previously unseen) ROIs, and the sheer enormity of the efforts needed to create a large pool of labels and atlases underscore the need for unsupervised segmentation methods. To illustrate this need, let us consider the problem of tumor detection from magnetic resonance (MR) images (see Figure~\ref{fig:1}(a)). On average each event detection involves the investigation of hundreds of slices that may vary significantly from patient to patient as well as over time. In such cases, retraining becomes inevitable, especially when novelty cases are encountered \cite{markou2006neural}. The spatial and temporal uncertainty in the morphology and location of ROIs further complicate these challenges. In addition, many a time data collection itself is extremely costly, e.g., imaging microscopic defects and microstructure anomalies in additively manufactured (AM) industrial components (see Figure~\ref{fig:1}(b) and Section 6.3 for details) tend to be time intensive. Together with cost and resource constraints, it becomes infeasible to gather big datasets for any given material system and processing recipe \cite{park2014estimating}.

Surprisingly, the literature on unsupervised segmentation is somewhat limited. The most notable methods include normalized cuts that is based on spectral graph partitioning \cite{shi2000normalized} and its subsequent derivatives such as watershed-based normalized cut \cite{yang2007improving} and multiscale normalized cut \cite{cour2005spectral}, mean shift clustering using the neighborhood pixel information \cite{comaniciu2002mean}, k-means clustering, expectation maximization and Markov random field-based methods \cite{zhang2001segmentation}. However, significant limitations exist, especially for the case of purely unsupervised approaches in terms of computational complexity, e.g., exact minimization of normalized cut is NP-complete \cite{shi2000normalized}, mean shift clustering requires manual specification of the kernel function and bandwidth \cite{nadler2007fundamental}.
 
Image segmentation approaches based on the energy minimization framework, particularly graph-based methods offer an elegant means to segment images without extensive training. Essentially, they involve minimizing an energy functional of the form \cite{kim2009color,boykov2001interactive,geman1984stochastic}:
\begin{equation}\label{eq:0}
E = E_{data}+E_{smooth}
\end{equation}
where $E_{data}$ accounts for the disagreement between the data samples, i.e., observed pixels and the estimated labels, and $E_{smooth}$ controls the smoothness of the labeling function. Continuous max-flow/min-cut algorithm is an efficient optimization approach to minimize the energy functions in polynomial time \cite{kolmogorov2004energy}. However, the problem of selecting the flow capacities prevents a fully unsupervised implementation. Most of the existing implementations of the continuous max-flow are either based on \textit{a priori} selection of the flow capacities \cite{appleton2006globally} or require manual interventions \cite{yuan2010continuous}. 

In this work, we present an approach to consistently estimating the flow capacities of a continuous max-flow/min-cut problem leading to fully unsupervised, fast image segmentation. Our framework is based on iteratively estimating the image labels by solving the max-flow problem while optimally estimating the flow capacities from the image characteristics. More specifically, we first setup the segmentation problem as a maximum \textit{a posteriori} estimation (MAP) of the image labels and show that it is equivalent to solving the max-flow problem given the flow capacities are known. By using the current optimum of the max-flow problem, we subsequently estimate the flow capacities by employing a novel Markov random field (MRF) prior over the flow capacities. In the sequel, we present theoretical results to establish the posterior consistency of the flow capacities based on the MRF prior proposed in this work. We implement our methodology on benchmark tumor datasets and a case study on the segmentation of defects in AM components that are fabricated using different processing recipes. An extensive comparison with state-of-the-art unsupervised algorithms suggests more than 90\% improvement in the average Dice score---a measure of the degree of overlap between the estimated labels and the ground truth (see Equation~\eqref{eq:22}). 

The remainder of the paper is organized as follows: In Section 2, we present a brief introduction to graph cuts for image segmentation in a continuous domain. In Section 3, we extend the graph cut to the max-flow problem in the continuous domain and show, via examples, that the estimation of flow capacities is critical to optimally solving the max-flow problem. In Section 4, we present our iterative MAP estimation approach to simultaneously estimate the maximum flow (i.e., the image labels) and the flow capacities. Section 5 presents theoretical results on the posterior consistency of the flow capacities. In section 6, we present comparative results for two distinct real-world case studies followed by conclusions in Section 7.

\begin{figure}
	\centering
	\includegraphics[width=0.45\textwidth]{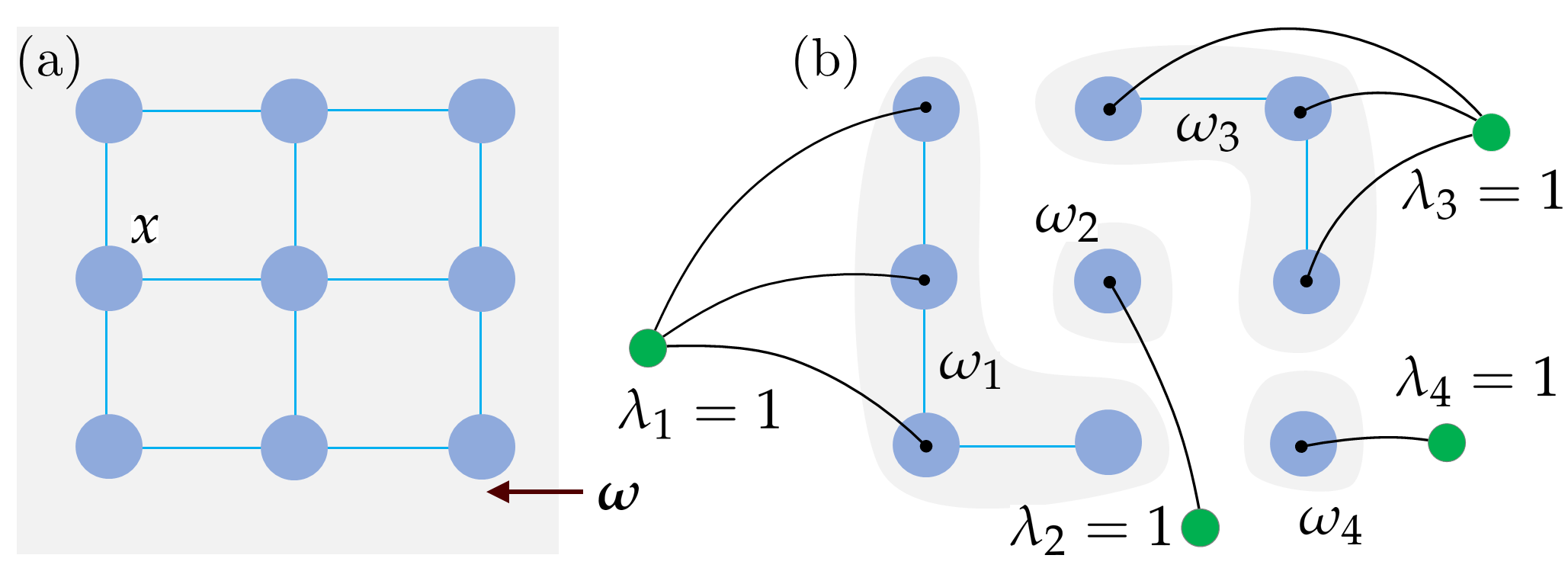}
	\caption{(a) {Graph representation} of an image $\bm{\omega}$ in a continuous domain. (b) A representative example of segmentation of the image $\bm{\omega}$ into subdomains $\omega_1,\ldots, \omega_k$ via graph cut.}
	\label{fig:1.1}
\end{figure}

\section{Image segmentation via graph cuts}
In this section, we begin with the basic notion of graph cuts. Let $\bm{\omega}$ be a continuous image domain (e.g., see Figure~\ref{fig:1.1}(a)) and $x$ be the sites (or pixels) in $\bm{\omega}$. The problem of image segmentation can be formulated as partitioning $\bm{\omega}$ into $n$ disjoint subdomains $\omega_1, \omega_2,\ldots, \omega_n$, with corresponding labeling function given as, 
\begin{equation}\label{eq:1}
\lambda_i(x)= 
\begin{cases}
1& x\in \omega_i \\
0 & x\not\in \omega_i \\
\end{cases}
\end{equation}
The partitioning is performed such that the resulting subdomains minimize the following energy functional \cite{chan2006algorithms}, 	
\begin{align}\label{eq:2}
\underset{\lambda_i(x) \in \{0,1\}}{\min} & \sum_{i=1}^{n}\int_{\bm{\omega}}\lbrace\lambda_i(x)\rho(x,\lambda_i(x)) +  C(x)|\nabla \lambda_i(x)|\rbrace dx 
\end{align}where the first term $\rho(x,\lambda_i(x))$ is the cost of assigning the site $x$ to the subdomain $\omega_i$ and is equivalent to the $E_{data}$ term (see  Equation~\eqref{eq:0}) when summed over the domain $\bm{\omega}$. The second term is the total variation (TV) regularization of $\lambda_i(x), \forall i\in\{1,2,\ldots,n\}$ where $C(x)>0$ controls the trade-off between the data term and the extent of regularization. TV regularization is particularly useful because of its property to selectively penalize the oscillations due to noise while preserving the discontinuities at the edges (see \cite{strong2003edge} for a mathematical justification).

 Note that the minimization function in Equation~\eqref{eq:2} is non-convex due to the binary configuration of $\lambda_i(x)$. To ensure tractability of the energy minimization function in Equation~\eqref{eq:2}, we consider the convex relaxation of $\lambda_i(x) = \{0,1\}$ to the unit interval $[0,1]$ as proposed by Chan et al. \cite{chan2006algorithms}. For simplification, we only consider the case of binary segmentation such that $n=2$, $\lambda(x) = 1-\lambda_1(x) = \lambda_2(x)$ (since $\sum_{i =1}^{n}\lambda_i(x) =1$), $C_s(x)= \rho(x,\lambda_1(x)) $ and $ C_t(x) = \rho(x,\lambda_2(x))$. Consequently, the energy minimization function for the case of binary label configuration can be written as,
\begin{align}\label{eq:3}
\underset{\lambda(x)\in[0,1]}{\min}  & \int_{\omega}(1-\lambda(x)) C_s(x) + \lambda(x) C_t(x) + C(x)|\nabla\lambda(x)|dx  
\end{align} 

Authors in \cite{chan2006algorithms} deduced that the optimal solution to Equation~\eqref{eq:2} can be obtained by thresholding the solution of the resulting convex problem in Equation~\eqref{eq:3}. However, the numerical algorithms for this minimization problem still suffer from the non-differentiability of the TV term $(\int_{\bm{\omega}}|\nabla \lambda(x)| dx)$ \cite{wei2016primal}. To overcome this, we investigate the dual of the energy function in Equation~\eqref{eq:3} and show that it is analogous to the continuous max-flow problem studied in \cite{yuan2010study,bae2011global}. 

\section{Continuous max-flow formulation}

In this section, we derive the dual of the energy function reported in Equation~\eqref{eq:3} and subsequently show that the dual is a max-flow problem in a continuous domain. We first consider the following results.
\begin{lemma}
Given $\lambda(x)$ is an indicator function as given in Equation~\eqref{eq:1}, it has bounded variation in $\bm{\omega}$, i.e., there exists a function $p(x)\in \mathcal{C}^1(\bm{\omega})$ with compact support and $|p(x)|\leq 1$ satisfying $\sup\lbrace\int_{\bm{\omega}} \lambda(x) \diver p(x) dx\rbrace<\infty$ such that, $$\int_{\bm{\omega}}\lambda(x) \diver p(x) dx = -\int_{\bm{\omega}}p(x) \cdot \nabla \lambda(x)  dx$$
\end{lemma}
\noindent See \cite{evans2018measure} for the proof. \qed
\begin{lemma} 
 With $p(x)$ defined as in Lemma 1, we have,
$$\int_{\bm{\omega}}C(x)|\nabla \lambda(x)| dx = \underset{|p(x)|\leq C(x)}{\max}\int_{\bm{\omega}}\lambda(x)\normalfont{\diver} p(x)dx  $$
\end{lemma}
\begin{proof*}
Using $p(x)$, the TV term in Equation~\eqref{eq:3} may be written as, 
$$ \int_{\bm{\omega}}C(x)|\nabla \lambda(x)| dx = \underset{|p(x)|\leq C(x)}{\max}\int_{\bm{\omega}} p(x)\cdot |\nabla\lambda(x)|dx$$From Lemma 1, we have, 
\begin{align*}
\int_{\bm{\omega}}\lambda(x)& \diver p(x) dx = -\int_{\bm{\omega}}p(x) \cdot \nabla \lambda(x)  dx \\ &\leq \int_{\omega}| p(x)\cdot\nabla\lambda(x)| dx \leq \int_{\omega}| p(x)|\cdot|\nabla\lambda(x)| dx
\end{align*}
Finally, taking the maximum over $|p(x)|$, we have,
\[
\pushQED{\qed} 
\int_{\bm{\omega}}C(x)|\nabla \lambda(x)| dx = \underset{|p(x)|\leq C(x)}{\max}\int_{\bm{\omega}}\lambda(x)\text{div}p(x)dx \qedhere
\popQED
\]     
\end{proof*}

\begin{figure}[!b]
	\centering
	\includegraphics[width=0.45\textwidth]{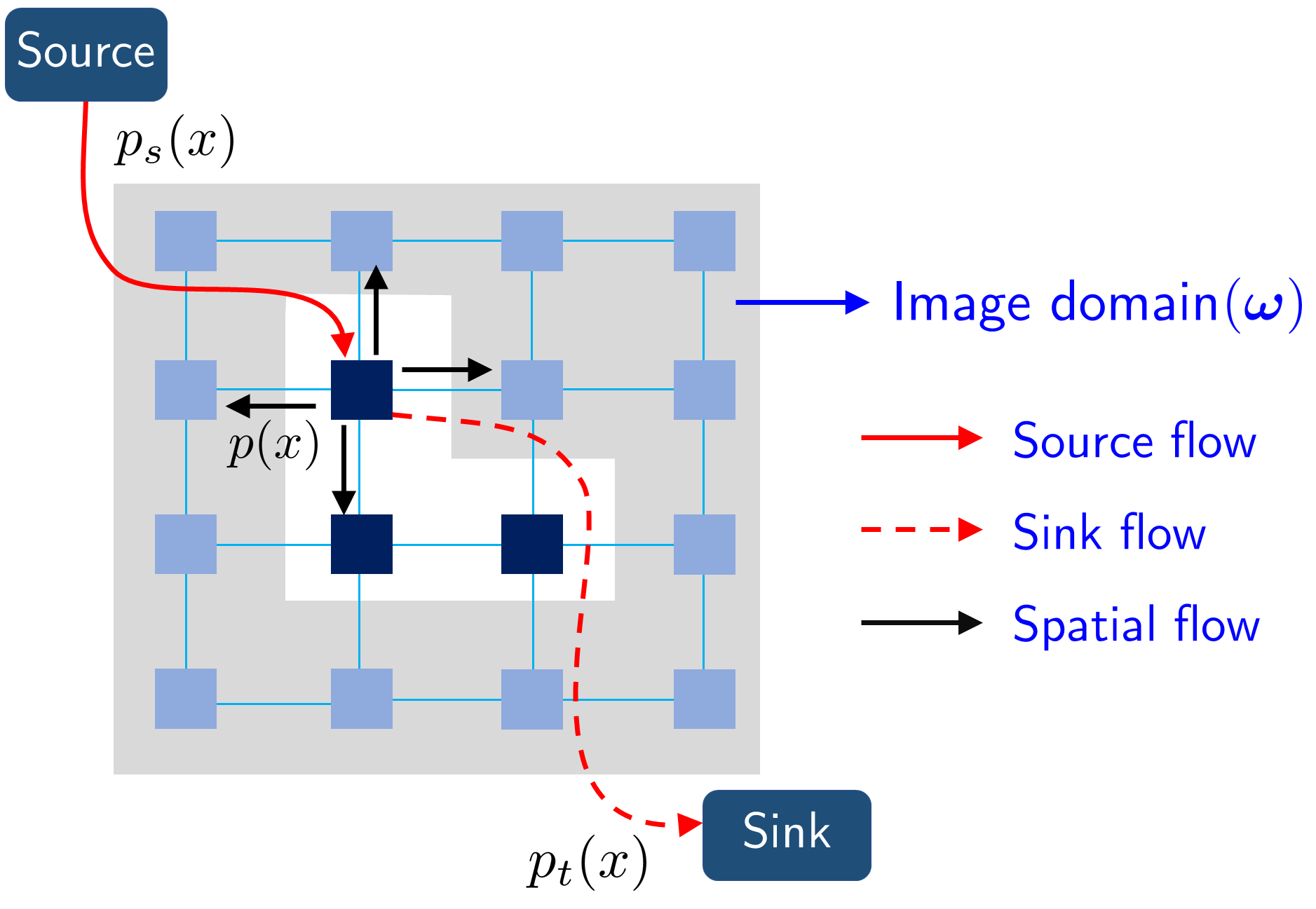}
	\caption{Representation of the continuous image domain where each of the sites are associated with source, sink and spatial flows represented by $p_s(x),p_t(x),p(x)$, respectively.}
	\label{fig:2}
\end{figure}

\begin{figure*}[!b]
	\centering
	\includegraphics[width=0.90\textwidth]{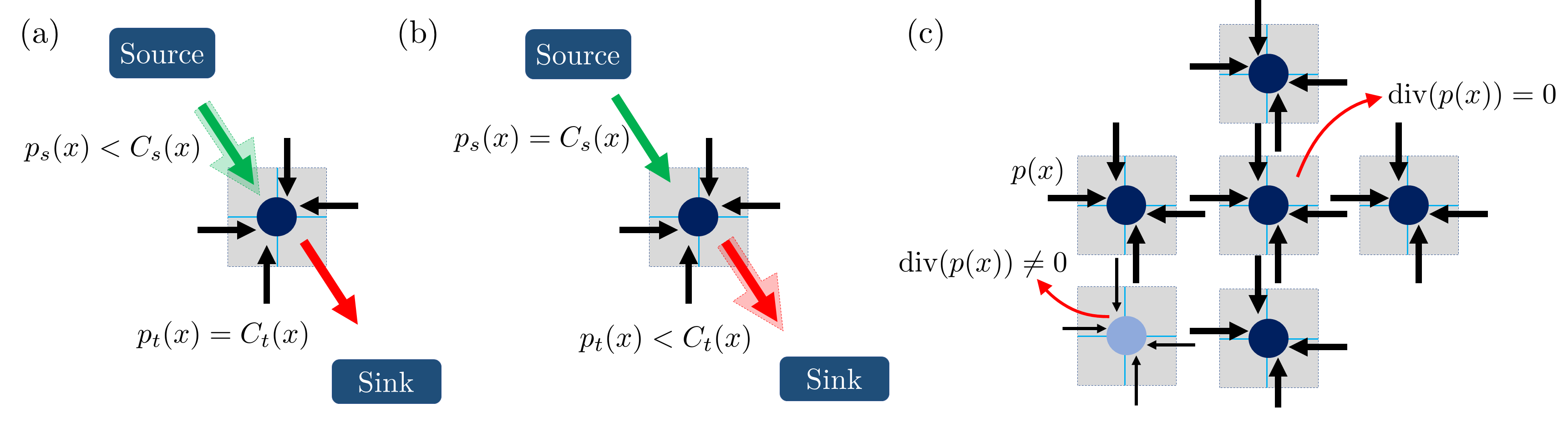}
	\caption{Visualization of the flow fields: (a) when the flow through $x$ is limited by the sink flow capacity such that the sink flow contributes to the total energy (b) when the flow through $x$ is limited by the source flow capacity such that the source flow contributes to the total energy and (c) shows two cases of divergence of the spatial flow. The spatial flow where $\text{div}p(x)\neq 0$ contributes to the total energy. We omit the edges connecting non-adjacent sites for the sake of clarity. }
	\label{fig:4-1}
\end{figure*}

Next, we consider functions $p_s(x)\leq C_s(x), p_t(x)\leq C_t(x)\in \mathbb{R}$ such that we get the following min-max equivalent of Equation~\eqref{eq:3}, 
\begin{align*}
\underset{\lambda(x)\in[0,1]}{\min}  \max_{\substack{p_s(x)\leq C_s(x)\\p_t(x)\leq C_t(x) \\ |p(x)|\leq C(x)}} \int_{\bm{\omega}}(1-\lambda(x))p_s(x) & + \lambda(x) p_t(x)  \\ & +  \lambda(x) \text{div}( p(x)) dx 
\end{align*}Note that the above min-max function is convex for fixed $\lambda(x)$ and concave for fixed $p_s(x), p_t(x)$, and $p(x)$ \cite{yuan2014spatially}. Via minimax theorem \cite{boyd2004convex}, we obtain the dual of Equation~\eqref{eq:3} as,
\begin{align}\label{eq:6}
\max & \int_{\bm{\omega}}p_s(x) dx  \\
\text{subject to }& ~ p_s(x)\leq C_s(x),~p_t(x)\leq C_t(x), ~|p(x)|\leq C(x) \nonumber\\
&\text{div} p(x) -p_s(x)  + p_t(x)  = 0~~\forall x\in\bm{\omega}\nonumber
\end{align}
The resulting formulation is the continuous analog of the discrete max-flow problem, i.e., maximizing the total source flow $p_s(x)$ subject to the source, sink and spatial flow capacities while satisfying the flow conservation principle.

To elaborate on the continuous max-flow problem, we begin by connecting each site $x$ to a source $s$ and a sink $t$ of infinite capacities. Each site in $\bm{\omega}$ is then associated with three different flow fields: the source flow $p_s(x)\in\mathbb{R}$, the sink flow $p_t(x)\in\mathbb{R}$, and the spatial flows $p(x)\in\mathbb{R}^2$. Here, the source and the sink flow fields are directed from the source $s$ to the site $x\in\bm{\omega}$ and from the site $x$ to the sink $t$, respectively. This is shown schematically in Figure~\ref{fig:2}. The spatial flow field is characterized by the undirected flow through $x$, thereby capturing the strength of interaction with the neighborhood locations. Each of the source, sink and spatial flows are constrained by their respective capacities represented as $C_s(x), C_t(x)$ and $C(x)$.

Returning to Equation~\eqref{eq:6}, we determine the maximum flow by writing the corresponding augmented Lagrangian as,
\begin{align}\label{eq:10}
\mathcal{L}(p_s,p_t,p,\lambda(x)) &=  \int_{\bm{\omega}}p_s(x)  + \lambda(x) (\text{div}( p(x)) -p_s(x) \nonumber \\  & \hskip-4em+ p_t(x) )dx  - \frac{c}{2}||\text{div}( p(x)) -p_s(x)  + p_t(x) ||^2
\end{align}
where $c>0$. See Algorithm 1 for solving the augmented Lagrangian $\mathcal{L}(p_s,p_t,p,\lambda(x))$ using the projection gradient descent approach \cite{yuan2010study}. 

The intuition behind the max-flow segmentation can be drawn from the idea of finding a cut with minimum capacity that will result in two disjoint partitions, one associated with the source and other with the sink. Let us consider a source flow $p_s(x)$ that is optimal but unsaturated, i.e., $p_s(x) \leq C_s(x)$ (see Figure~\ref{fig:4-1}(a) for reference). Since the optimal source flow is unsaturated, it has no contribution to the total energy. As a result, any variation in the source flow should not change the total energy and therefore, $1-\lambda(x)$ must be equal to zero, or  $\lambda(x)=1$. This implies that $p_t(x) = C_t(x)$, i.e., the flow from site $x$ to sink $t$ is saturated and the minimum cut passes through this edge. Similarly, the unsaturated sink flows, i.e., $p_t(x) \leq C_t(x)$ lead to saturated flows from source $s$ to the site $x$ as shown in Figure~\ref{fig:4-1}(b). For this case, the minimum cut severs the source flow and assigns the site $x$ to the sink (i.e, $\lambda(x)=0$). Finally, for the case of spatial flows, the sites for which $\text{div}p(x) \neq 0$ are the candidates for minimum cut. In other words, the maximum flow occurs through the sites that are saturated, and therefore, are the candidates for minimum cut (see Figure~\ref{fig:4-1}(c)). 
\begin{algorithm}[!t]
	\SetKwInOut{Input}{Initialize}
	\SetKwInOut{Output}{Output}
	\Input{$p^0_s(x),p^0_t(x),p^0(x),\lambda^0$, $k=0$ \& step size $\gamma$} 
	\Repeat{convergence}{
			{\color{gray}\%\textit{ update the spatial flows}}\\
			$D^k(x) \leftarrow p^k_s(x) + \lambda^k(x)/c - p^k_t(x)$\;
			$p^{k+1}(x) \leftarrow p^{k}(x) -\gamma\nabla(\text{div}p^k(x) - D^k(x))$ \;
			{\color{gray}\%\textit{ update the sink flows}}\\
			$F^k(x) \leftarrow p^k_s(x) + \lambda^k(x)/c - \diver p^{k+1}(x)$\;
			$p_t^{k+1}(x) \leftarrow \min (C_t(x) ,F^k(x) )$\;	
			{\color{gray}\%\textit{ update the source flows}}\\
			$G^k(x) \leftarrow p^{k+1}_t(x) - \lambda^k(x) + \diver p^{k+1}_t(x)/c$\;
			$p_s^{k+1}(x) \leftarrow (1+cG^k(x))/2c$\;
			{\color{gray}\%\textit{ update the multipliers}}\;
			$\epsilon^{k+1} \leftarrow \text{div} p^{k+1}(x) -p^{k+1}_s(x)  + p^{k+1}_t(x)$\;
			$\lambda^{k+1}(x) \leftarrow \lambda^k(x) - c\epsilon^{k+1}$\;
		}
	\caption{Projection gradient descent}
\end{algorithm}

\begin{figure*}[t]
	\centering
	\includegraphics[width=0.80\textwidth]{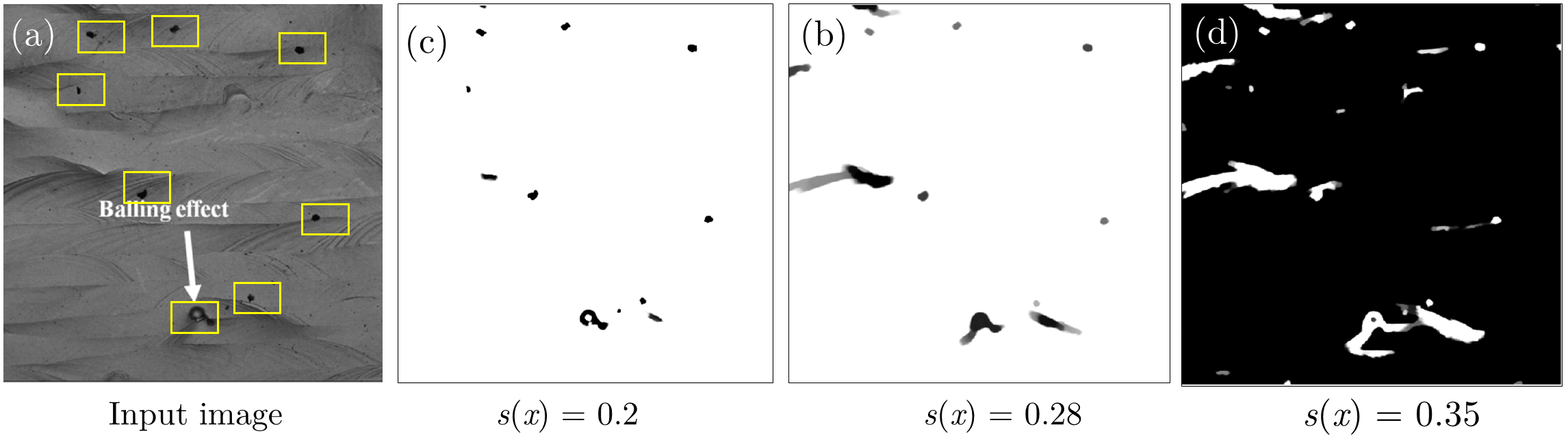}
	\caption{ Representative example to illustrate the effect of variations in the source flow capacity on the segmentation result. (a) Input image showing the defects. Segmentation using the continuous max-flow approach with $t(x)$ and spatial flow capacities fixed to 0.3 and 0.1 respectively, but with varying values of $s(x)$. (b) $s(x) = 0.2$, (c) $s(x) = 0.28$ and (d) $s(x) = 0.35$.}
	\label{fig:3}
\end{figure*}

However, difficulties arise when estimating the flow capacities. Generally the flow capacities are defined as $C_s(x) = D(I(x) - s(x)), C_t(x) = D(I(x)-t(x))$ where $I(x)$ is the image matrix containing pixel intensities, $D(.)$ is some user defined function and $s(x)$ and $t(x)$ are selected \textit{a priori} \cite{yuan2010study} or defined heuristically \cite{appleton2006globally}. Nonetheless, the flow capacities are generally unknown and require trial and error or user inputs in the form of bounding boxes or scribbles for estimation. To argue the importance of flow capacities in optimally solving the max-flow problem, we begin with a trivial case of uniformly constant source and sink flow capacities. In such a scenario, sites have no preference to be assigned to the source or the sink and consequently, all the sites are either assigned to the source or the sink depending on which flow field is saturated first. Appleton and Talbot \cite{appleton2006globally} addressed this problem by defining a weighting function that decays as per the power-law in the neighborhood of user defined initialization or seed points. Alternatively, Boykov and Kolmogorov \cite{boykov2003computing} used the Cauchy-Crofton formula to arrive at the (edge) capacities. However, both of these approaches required manual intervention to  define the seed points. 

For cases where flow capacities are spatially varying, the segmentation result may still vary significantly if the values are not optimally chosen. To demonstrate this, we consider the problem of identifying defects (marked with yellow boxes in Figure~\ref{fig:3}(a)) on an additively manufactured component (see Section 6.3 for additional details). To define the flow capacities, we use $D(.)\equiv |.|$ and $t(x) = 0.3$. Figures~\ref{fig:3}(b-d) show the segmentation results for three different values of $s(x)$. We note that even small perturbations in the value of $s(x)$ result in vastly different segmentation; demonstrating the need for estimating the flow capacities to optimally solve the max-flow problem.

\section{Estimation of flow field capacities} 
In this section, we develop an approach to estimate the image labels $\lambda(x)$ by solving the continuous max-flow problem while optimally estimating the flow capacities from the image characteristics. Towards this, we first formulate the problem of estimating $\lambda(x)$ as a MAP estimation problem. Given the image $I(x)$ and the flow capacities $\mathcal{C}(x) = \{C_s(x), C_t(x), C(x)\}$, the objective of the MAP estimation is to find the image labels ${\lambda(x)}$ that maximizes the posterior probability given as,
\begin{equation}\label{eq:11}
{\lambda(x)} = \arg\underset{\lambda(x)}{\max }~P(\lambda(x)| \mathcal{C}(x),I(x))
\end{equation}
Since the flow capacities are unknown, we estimate the image labels $\lambda(x)$ in Equation~\eqref{eq:11} by iteratively solving the following two subproblems,
\begin{subequations}
\begin{equation}\label{eq:12a}
{\lambda}^{\tau+1}(x) = \arg\underset{\lambda(x)}{\max }~P(\lambda(x)|\mathcal{C}^{\tau}(x),I(x)) 
\end{equation}
\begin{equation}\label{eq:12b}
\mathcal{C}^{\tau+1}(x) = \arg\underset{\mathcal{C}(x)}{\max }~P(\lambda^{\tau+1}(x)|\mathcal{C}(x), I(x)) 
\end{equation}
\end{subequations}
where $\lambda^{\tau+1}(x)$ and $\mathcal{C}^{\tau+1}(x)$ are respectively the image labels and the flow capacities at iteration $\tau+1$. The first subproblem (Equation~\eqref{eq:12a}) is essentially maximizing the posterior distribution of the image labels given the image $I(x)$ and the most recent estimate of flow capacities, i.e., $\mathcal{C}^{\tau}(x)$. In the second subproblem (Equation~\eqref{eq:12b}), we employ the updated estimate of $\lambda^{\tau+1}(x)$ to subsequently update the flow capacities. The process is repeated until a stopping criterion (or convergence) is reached.

\subsection{Maximum {a posteriori} estimation of $\lambda(x)$}

We now employ a MAP estimation approach to determine the optimal image labels $\lambda(x)$ by solving the first subproblem as given in Equation~\eqref{eq:12a}. Using Bayes theorem we have, 
\begin{align}\label{eq:15}
\lambda^{\tau+1}(x) =  \arg\underset{\lambda(x)}{\max }\log(P(I(x)|{\lambda(x)},\mathcal{C}^{\tau}(x)))+\log(P({\lambda(x)})
\end{align}
Here, the first term is the log-likelihood of the image labels ${\lambda}(x)$ given the flow capacities and the image $I(x)$. In other words, it imposes a penalty for every incorrect assignment of the labels. Assuming the observations (i.e., pixel intensities contained in $I(x)$) are independently and identically distributed we can write,  
\begin{equation}\label{eq:16a}
P(I(x)|{\lambda(x)},\mathcal{C}^{\tau}) \propto \exp(-D(\mathcal{C}^{\tau},\lambda(x),I(x)))
\end{equation}where $D(\mathcal{C}^{\tau},\lambda(x),I(x))$ is the data penalty function defined as,  
\begin{equation*}
D(\mathcal{C}^{\tau},\lambda(x),I(x)) = \int_{\bm{\omega}}(1-\lambda(x))C^{\tau}_s(x)  + \lambda(x) C^{\tau}_t(x) dx
\end{equation*}The second term in Equation~\eqref{eq:15} is the logarithm of the prior over flow capacities. To enforce spatial smoothness, we define $P({\lambda(x)})$ as, 
\begin{equation}\label{eq:16b}
P({\lambda(x)}) \propto \exp\left(-V(C^{\tau}(x), \lambda(x))\right)
\end{equation}where $C^{\tau}(x)$ is the estimate of spatial flow capacity at iteration $\tau$ and $V(C^{\tau}(x), \lambda(x))$ is the smoothness penalty given as, $$V(C^{\tau}(x), \lambda(x)) =C^{\tau}(x)\int_{\bm{\omega}}|\nabla \lambda(x)| dx$$ Substituting the values of $P(I(x)|\lambda(x), \mathcal{C}^{\tau})$ and $P(\lambda(x))$ in Equation~\eqref{eq:15}, we have, 
\begin{align*}\label{eq:17}
\lambda^{\tau+1}(x) =  \arg\underset{\lambda(x)}{\min}~ \int_{\bm{\omega}}(1-\lambda(x))C^{\tau}_s(x) &+ \lambda(x) C^{\tau}_t(x) \\  & + C^{\tau}(x)|\nabla \lambda(x)| dx \nonumber
\end{align*}

We note that the maximization of the posterior probability in Equation~\eqref{eq:12a} is equivalent to solving the max-flow problem given the flow capacities are known. We, therefore, simply use Algorithm 1 to find the optimal values of ${\lambda^{\tau+1}(x)},p^{\tau+1}_s(x),p^{\tau+1}_t(x)$ and $p^{\tau+1}(x)$ at iteration $\tau+1$.

\subsection{Maximum {a posteriori} estimation of $\mathcal{C}$}

We now look at the second subproblem, i.e., 
\begin{equation}\label{eq:13}
\mathcal{C}^{\tau+1}(x) =  \arg\underset{\mathcal{C}(x)}{\max } \log{P({\lambda^{\tau+1}(x)}|\mathcal{C}(x), I(x))}+ \log P(\mathcal{C}(x))
\end{equation}
Here, the first term is the log-likelihood of the flow capacities $\mathcal{C}(x)$ given the image $I(x)$ and the most recent estimate of the image labels $\lambda^{\tau+1}(x)$ and $P(\mathcal{C}(x))$ is the prior distribution over the flow capacities. Note that the prior over the flow capacities is critical to enforce spatial smoothness and optimally solving the max-flow problem as discussed in Section~3. Several approaches have been proposed in the literature to capture the spatial smoothness, most commonly using MRF priors \cite{boykov1998markov, geman1984stochastic, derin1987modeling}. {\color{black}Using the prior presented in \cite{nguyen2013fast}, we propose a new MRF prior that is computationally fast and enforces spatial smoothness. For the simplicity of notations, we only focus on the estimation of the source flow capacity given as $C_s(x) = |I(x) - s(x)|$. Results may be generalized for estimating the sink and spatial flow capacities similarly. }

\subsubsection{Prior over the flow fields}

MRF priors have been widely used to capture the spatial smoothness in segmentation problems. One of the most commonly used families of MRF priors is given as, 
\begin{equation}\label{eq:13a}
p(s(x)) = \frac{1}{Z}\exp\left(-\frac{1}{T}U(s(x))\right)
\end{equation}
where $Z$ is a normalizing constant, $T$ is a scale factor and $U(s(x))$ is a smoothing function that controls the spatial correlation between the sites $x\in\bm{\omega}$ in a given neighborhood \cite{li2009markov}. The prior follows from the Hammersley-Clifford Theorem \cite{grimmett1973theorem} when assuming the local Markovian property over the flow field, that is, $p(s(x)|\bm{\omega}\backslash {x})) = p(s(x)|x\in \mathcal{N}(x))$ where $\mathcal{N}(x)$ is the . It may be noted that the local Markovian property is quite natural for random fields defined over an image domain \cite{boykov1998markov}.

Several choices of the smoothing function have been proposed in the literature, e.g., see \cite{li2009markov,besag1974spatial}. However, the complexity of the posterior computation remains one of the critical challenges under MRF priors. To minimize the computational complexity, we present the following smoothing function based on \cite{nguyen2013fast}, 
\begin{equation}\label{eq:14b}
U(s^{\tau+1}(x))  = -\sum_{x\in\bm{\omega}}\left(G_s^{\tau}(x) + \beta(1-p^{\tau}_s(x))\right)\log \left( s^{\tau+1}(x)\right)
\end{equation}
where the multiplier $G_s^{\tau}(x)$ is defined as, 
\begin{equation}\label{eq:14c}
G_s^{\tau}(x) = \exp\left(\frac{\beta}{2|\mathcal{N}(x)|}\sum_{y\in\mathcal{N}(x)}(p^{\tau}_s(y)+s^{\tau}(y)) \right)
\end{equation}

Here, $|\mathcal{N}(x)|$ is the cardinality of the neighborhood of $x$, $p_s^{\tau}(x)$ and $s^{\tau}(x)$ are the current estimates of the source flow $p_s(x)$ and $s(x)$, respectively, and $\beta$ is a smoothing constant. The term $G_s^{\tau}(x)$ in Equation~\eqref{eq:14b} enforces spatial smoothness by mean filtering over the neighborhood $\mathcal{N}(x)$. This minimizes the slack between the flow field and flow capacity. {\color{black} Although the smoothing function is heuristically designed, an inherent advantage is that when maximizing the log-likelihood function the derivative is dependent only on the term $s^{\tau+1}(x)$ at the step $\tau+1$, thereby contributing to the computational efficiency. In the present implementations, we set the value of $\beta=5$ and the neighborhood size as $5\times5$ over $\bm{\omega}$ such that $|\mathcal{N}(x)|= 25$. See \cite{nguyen2013fast} for details on the parameter selection. }

\subsubsection{Updating the flow capacities}

We now determine the value of the source flow capacities $C^{\tau+1}_s(x)$ by maximizing the posterior probability in Equation~\eqref{eq:13}. With respect to $s^{\tau+1}(x)$, we have, 
\begin{equation*}
\frac{d}{ds^{\tau+1}(x)}\left(\log{P(I(x)|\mathcal{C}(x),{\lambda^{\tau+1}(x)})}+ \log P(\mathcal{C}(x))\right) = 0
\end{equation*}
For the first term, we refer to Equation~\eqref{eq:16a} and consider the dual of the corresponding minimization problem stated in Equation~\eqref{eq:6}. Using Lagrange multipliers $\eta(x)$ for the source constraints, we have, 
\begin{align*}
&\frac{d}{ds^{\tau+1}(x)} \bigg(\max \int_{\bm{\omega}}p_s(x) dx -\frac{1}{T} \sum_{x\in\bm{\omega}}(G_s^{\tau}(x)  +  \beta(1-p^{\tau}_s(x)))  \nonumber \\ &\times\log \left( s^{\tau+1}(x)\right) -  \sum_{x} \eta(x)(p_s(x)- |I(x)- s^{\tau+1}(x)|)\bigg) =0
\end{align*}On simplification, this gives, $$s^{\tau+1}(x) = \frac{1}{T}\frac{G_s^{\tau}(x)  +  \beta(1-p^{\tau}_s(x))}{\eta(x)} $$For simplicity, we use $T = \eta(x)^{-1}$ such that the source flow capacity in iteration $\tau+1$ is given as,  
\begin{equation}\label{eq:21}
C^{(\tau+1)}_s(x) = |I(x) - \left(G_s^{\tau}(x) + \beta(1-p_s^{\tau}(x))\right)|
\end{equation} For the sink flow capacity $(C_t(x) = |I(x) - t(x)|$, we define the smoothing function for $t(x)$ similar to Equation~\eqref{eq:14b} as, 
\begin{equation*}\label{eq:14d}
U(t^{\tau+1}(x))  = -\sum_{x\in\bm{\omega}}\left(G_t^{\tau}(x) + \beta p^{\tau}_t(x)\right)\log \left( t^{\tau+1}(x)\right)
\end{equation*}such that the sink flow capacity at iteration $\tau+1$ is, 
\begin{equation}\label{eq:21b}
C^{(\tau+1)}_t(x) = |I(x) - \left(G_t^{\tau}(x) + \beta p_t^{\tau}(x)\right)|
\end{equation}where 
\begin{equation}\label{eq:21d}
G_t^{\tau}(x) = \exp\left(\frac{\beta}{2|\mathcal{N}(x)|}\sum_{y\in\mathcal{N}(x)}(p^{\tau}_t(y)+s^{\tau}(y)) \right)
\end{equation}
\noindent Finally, to update the spatial flow capacities, we use the flow conservation constraint in Equation~\eqref{eq:6} as, 
\begin{equation}\label{eq:21c}
C^{\tau+1}(x) = |\int_{\bm{\omega}} C^{\tau+1}_s(x)dx - \int_{\bm{\omega}}C^{\tau+1}_t(x)dx| 
\end{equation}
The algorithm for simultaneous update of all the flow capacities is given in Algorithm 2.

\begin{algorithm}[!t]	
	\SetKwInOut{Input}{Initialize}
	\SetKwInOut{Output}{Output}
	\Input{Initialize the parameters ${\lambda^{(0)}},p^{(0)}_s(x),p^{(0)}_t(x),p^{(0)}(x)$ and $\mathcal{C}^{(0)}(x)$ } 
	\Repeat{convergence}{
		Solve the continuous max-flow problem in Equation~\eqref{eq:10} to estimate ${\lambda^{\tau}(x)},p^{\tau}_s(x),p^{\tau}_t(x)$ and $p^{\tau}(x)$\;
		Evaluate the current estimate of $G_s^{\tau+1}(x)$ in Equation~\eqref{eq:14c} and $G_t^{\tau+1}(x)$ in Equation~\eqref{eq:21d}\;
		Update the flow capacities as:\\
		 $	C^{(\tau+1)}_s(x) = |I(x) - \left(G_s^{\tau}(x) + \beta(1-p_s^{\tau}(x))\right)|$ \\
		 $	C^{(\tau+1)}_t(x) = |I(x) - \left(G_t^{\tau}(x) + \beta p_t^{\tau}(x)\right)|$ \\
		 $	C^{\tau+1}(x) = |\int_{\bm{\omega}} C^{\tau+1}_s(x)dx - \int_{\bm{\omega}}C^{\tau+1}_t(x)dx|$	
	}
	\caption{Iterative MAP estimate of source, sink and spatial capacities}
\end{algorithm}

\section{Convergence and consistency}

Inconsistency of the posterior is one of the biggest challenges while working with heuristic or otherwise non-standard priors as noted by Freedman \cite{freedman1965asymptotic}: \textit{the set of all parameter-prior pairs $(\theta,p)\in \Theta\times\mathcal{P}$ for which the posterior is consistent at $\theta$ is a meager set.} Inconsistency is more common than expected and may lead to incorrect inferences. To avoid situations like this and others as indicated in \cite{barron1999consistency}, we show that the posterior probability of the flow capacities based on the MRF prior with smoothing function given in Equation~\eqref{eq:14b} is consistent. 

Let us consider the set of independently and identically distributed image observations $I(x)$ taking values in the measurable space $(\Omega,\mathcal{B})$ and are sampled from some unknown ``true'' distribution function $F_0$ with density function $f_0$. Given the observations $I(x)$, let $p(s(x)|I(x))$ denote the posterior distribution of $s(x)$. Let $\mu$ be a probability measure on the measurable space $(\Omega,\mathcal{B})$. We define a $\varepsilon$-Hellinger neighborhood of the true distribution $F_0$ as $s_{\varepsilon}(x)= \{F\in \mathcal{P} :{H}(F_0,F)\leq\varepsilon\} $  where ${H}(F_0,F)$ denotes the Hellinger distance and is given as, $${H}(F_0,F) = \left\{\left(\int \sqrt {f_0}-\sqrt {f} \right)^2\mu(d\omega)\right\}^{1/2} $$and $\mathcal{P}$ is the subset of all finite probability measures that are absolutely continuous with respect to $\mu$. Then the consistency of the posterior distribution relies on the following two assumptions \cite{barron1999consistency},

\begin{assumption}
For every $\varepsilon>0$, $p(N_{\varepsilon})>0$ where $N_{\varepsilon}$ is the $\varepsilon$-Kullback-Leibler neighborhood \cite{barron1999consistency} of $F_0$. 
\end{assumption}
\begin{assumption}
For every $\varepsilon>0$, there exists a sequence $\{\mathcal{F}_{n=1}^{\infty}\}$ of subsets of $\mathcal{P}$, and positive real numbers $c,c_1,c_2, \delta$ with $c<([\varepsilon-\sqrt{\delta}]^2-\delta)/2, ~\delta < \varepsilon^2/4$, such that: 
\begin{enumerate}[(i)]
	\item $p(\mathcal{F}^c_n)\leq c_1\exp(-nc_2)$ for all but finitely many $n$, and
	\item $\mathcal{H}(\mathcal{F}_n,\delta)\leq nc$ for all but finitely many $n$ where $\mathcal{H}$ is the $\delta$-metric entropy of $\mathcal{F}_n$ (see Definition 1 in \cite{barron1999consistency}). 
\end{enumerate}
\end{assumption}

Intuitively, Assumption 1 is needed to ensure positive prior probability in the neighborhood of $F_0$. The second assumption prevents the prior from giving substantial mass to distributions with ``wiggly'' densities (condition (\textit{i})). This is realized by imposing exponentially small probability to the complement of the sequence $\mathcal{F}_n$ of well behaved densities, i.e., not wiggly. Condition (\textit{ii}) ensures that the sequence $\mathcal{F}_n$ of densities are well-behaved. Under these assumptions, \cite{barron1999consistency} showed that the posterior distribution is consistent at $F_0$, i.e., the posterior concentrates all the mass in the $\varepsilon$-Hellinger neighborhood almost surely with probability 1. More formally, we have,

\begin{theorem}
The posterior concentrates all the mass in the $\varepsilon$-neighborhood of $F_0$ almost surely, i.e., $p(s_{\varepsilon}(x)|I(x))\rightarrow1$ almost surely as $n\rightarrow\infty$.
\end{theorem}

\noindent To show that the result holds, we first present the following lemma. 

\begin{lemma}
The MRF prior $p(s(x))$ in Equation~\eqref{eq:13a} belongs to an exponential family.
\end{lemma}

\begin{proof*}
Plugging the values of $U(s(x))$ and $G(x)$ from Equations~\eqref{eq:14b} and~\eqref{eq:14c}, respectively in Equation~\eqref{eq:13a}, we have: 
$$p(s(x)) = \frac{1}{Z}\exp\left(\frac{1}{T}{\underset{\bm{\omega}}{\sum} [e^{c_3(\tilde{p}_s+\tilde{s})} + \beta(1-p_s(x))]\log(s(x))}\right)$$where $c_3 = \beta/2\mathcal{N}(x)$, $\tilde{p}_s = \underset{y\in\mathcal{N}(x)}{\sum}p_s(y)$, and $\tilde{s} = \underset{y\in\mathcal{N}(x)}{\sum}s(y)$. Factorizing the parameter and variable terms, we rewrite $p(s(x))$ as,
\begin{align}\label{eq:12a}
p(s(x)) =&\exp\left(\underset{\bm{\omega}}{\sum} [ e^{c_3\tilde{p}_s}\log(s(x)) + \beta\log(s(x))] \right)\times \nonumber\\ &\exp\left(-\underset{\bm{\omega}}{\sum} [\beta p_s(x) \log(s(x))-e^{c_3\tilde{s}}\log(s(x))]\right)\nonumber
\end{align}
Since the product of exponential families belongs to the exponential family, $p(s(x))$ belongs to the exponential family. \qed
\end{proof*}
  
 Given the Assumptions 1 and 2 and the prior $p(s(x))$ belonging to the exponential family, the posterior consistency and therefore, the proof of Theorem 1 follows from \cite{barron1999consistency}. In the next section, we present the implementation results and comparison with state-of-the-art segmentation methods.


\section{Experimental results}

\subsection{Case studies and evaluation metrics}
\begin{figure*}[!t]
	\centering
	\includegraphics[width=0.85\textwidth]{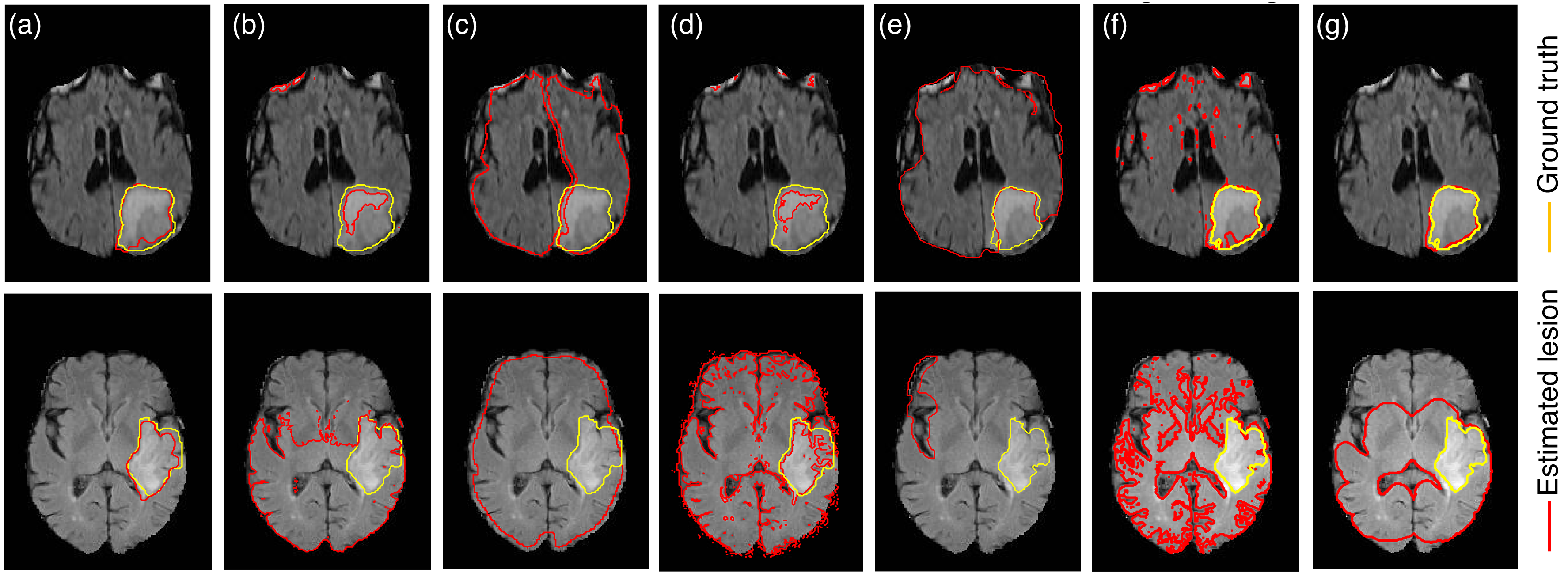}
	\caption{Segmentation results of different unsupervised approaches for the segmentation of brain tumor on HG (top row) and LG (bottom row) glioma. (a) The proposed method (b) mean shift (c) normalized cuts (d) blobworld (e) hierarchical image segmentation (f) GMM and (g) spatially constrained GMM.}
	\label{fig:6.1}
\end{figure*}

To test the efficacy of the proposed methodology, we investigate two real-world case studies: Brain Tumor Segmentation (BraTS) using MR images and defect segmentation in additively manufactured components using scanning electron microscope (SEM) images. While BraTS dataset is sufficiently large to enable supervised segmentation of lesions (tumorous regions), significant uncertainties exist in terms of morphology, location, and intensity values. These challenges render on the fly segmentation of lesions extremely difficult. In contrast, the second case study involves high-resolution SEM images that are extremely costly to record. In such scenarios, limited datasets hinder the application of supervised segmentation methods. Nonetheless, both the case studies call for an unsupervised approach to enable on the fly segmentation of ROIs.

Both supervised as well as unsupervised segmentation approaches have been reported in the literature with the majority being supervised. Therefore, we compare our segmentation results with both supervised as well as unsupervised algorithms. For the first case study, we refer to the supervised algorithms reported in the Medical Image Computing and Computer Assisted Intervention Society (MICCAI) proceedings 2013 and 2015 \cite{menze2015multimodal}. In the 2013 MICCAI proceedings, 20 algorithms were reported that include five generative, 13 discriminative and two generative-discriminative algorithms. In the 2015 BRATS dataset, three generative, two discriminative and seven neural network/deep learning approaches were reported. For unsupervised segmentation, we refer to k-means, expectation maximization (also referred to as blobworld) \cite{carson2002blobworld}, Gaussian mixture model (GMM), GMM with spatial regularization (SC-GMM) \cite{nguyen2013fast}, mean shift \cite{comaniciu2002mean}, normalized cuts \cite{shi2000normalized}, and hierarchical image segmentation \cite{arbelaez2011contour}. Due to the limited dataset in the second case study, we compare the performance with only unsupervised algorithms.  

\begin{figure}[!t]
\centering
\includegraphics[width = 0.40\textwidth]{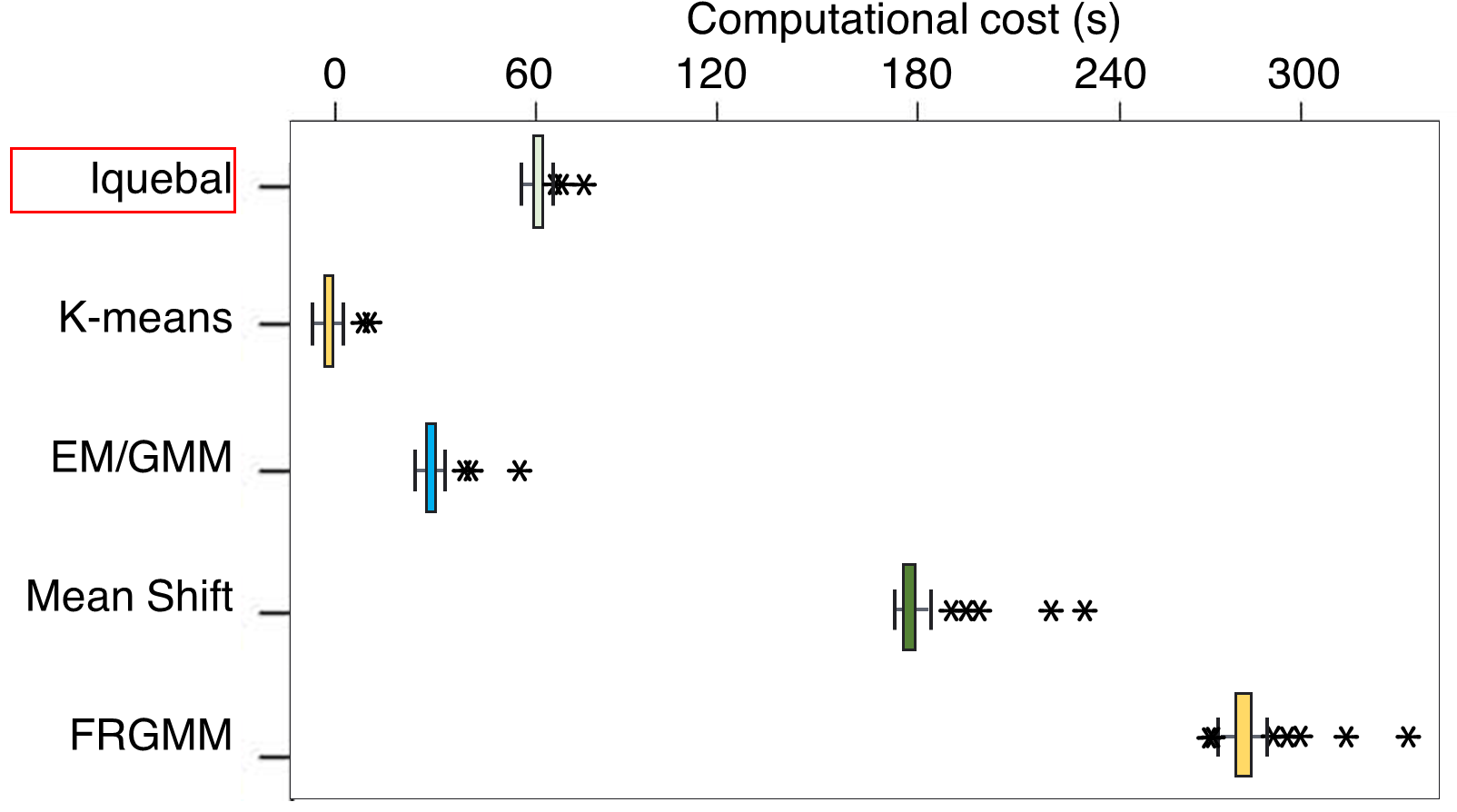}
\caption{Comparison of the computational cost among the unsupervised algorithms that converged.}
\label{fig:7.1}
\end{figure}

\begin{figure*}[!t]
	\centering
	\includegraphics[width=0.80\textwidth]{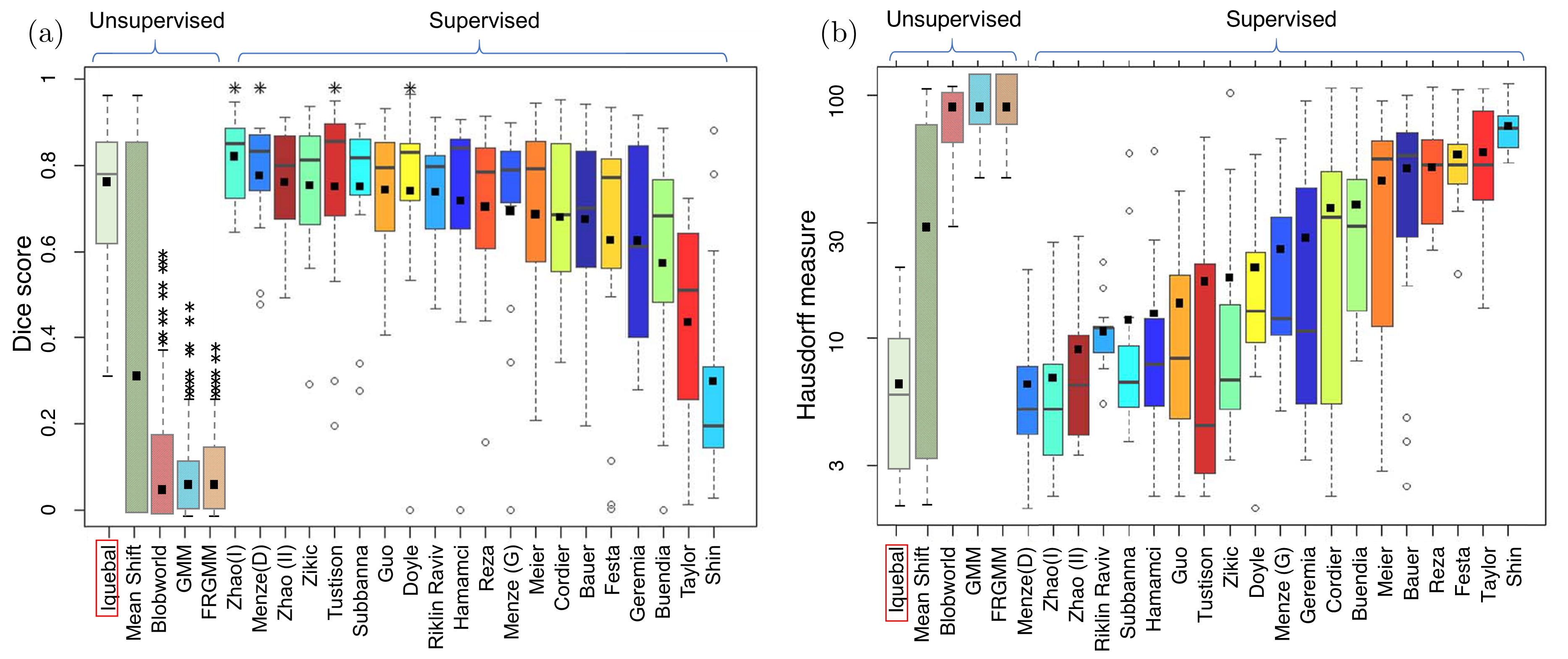}
	\caption{Comparative results of different algorithms tested for the segmentation of brain tumor on the BRATS 2013 dataset. Box plot adopted from \cite{menze2015multimodal}}
	\label{fig:8.1}
\end{figure*}

To quantitatively compare the performance of the segmentation results, we refer to the standard Dice score and the Hausdorff distance. The Dice score is given as, \begin{equation}\label{eq:22}
\text{Dice}(\hat{\mathcal{R}},\mathcal{R}) = 2\frac{|\hat{\mathcal{R}}\cap\mathcal{R}|}{|\hat{\mathcal{R}}+\mathcal{R}|}
\end{equation}
where $\hat{\mathcal{R}}$ and $\mathcal{R}$ represent the segmented lesion/defects and the expert segmentation, respectively, and $|.|$ represents the size of the domain. Dice score measures the areal overlap or the agreement between the segmented area and the ground truth. The Hausdorff distance  instead measures the surface distance between the segmented area and the ground truth and is given as, 
\begin{equation}\label{eq:23}
\text{Haus}(\hat{\mathcal{R}},\mathcal{R}) = \max\{\adjustlimits\sup_{i\in \hat{\mathcal{R}}}  d(i,\mathcal{R}), \adjustlimits\sup_{j\in {\mathcal{R}}} d(j,\hat{\mathcal{R}})\}
\end{equation}
where $d(j,\hat{\mathcal{R}})$ is the shortest Euclidean distance between the sites in $\mathcal{R}$ and $\hat{\mathcal{R}}$. Taking maximum over the supremum of these Euclidean distances make Hausdorff distance highly sensitive to the outliers present in $\hat{\mathcal{R}}$. To control this, we use the 95th percentile of the Hausdorff distance as suggested in \cite{menze2015multimodal}.

\subsection{Brain Tumor Segmentation}


Despite significant advances over the past few years, the diagnosis of glioma---the most common type of brain tumor---remains limited. Neuroimaging offers a noninvasive approach to evaluate the progression of the lesions, thus allowing timely intervention and controlled treatment. Lesions are typically imaged by using fluid-attenuated inversion recovery (FLAIR) MR scans that highlights the differences in water relaxation properties of the tissues, thereby enhancing the peritumoral edema---a characteristic feature of malignant glioma. As the lesions are defined only in terms of the contrast changes, the boundaries are often ill-defined and lead to variations among the expert segmentation \cite{menze2015multimodal}. Lesions may also vary greatly in terms of size, shape, and localization from patient to patient, as well as across subsequent stages of the tumor growth, thereby requiring large training datasets for implementing a supervised segmentation approach. 

In this study, we use the publicly available FLAIR-MR scans from BraTS 2013 and 2015 challenge. There are a total of 20 high grade (HG) and ten low grade (LG) glioma cases in the BraTS 2013 dataset and 274 HG and 54 LG glioma in the BraTS 2015 dataset. Each of these datasets contains annotated ground truth, already delineated by the clinical experts. All the images were resampled to the size of $240 \times 240$.

Segmentation results of a representative HG and LG glioma are presented in the top and bottom rows of Figure~\ref{fig:6.1}, respectively. The segmentation results for the present approach is shown in Figure~\ref{fig:6.1}(a) followed by rest of the unsupervised segmentation methods in Figures~\ref{fig:6.1}(b)-(g). Note that some of the methods such as mean shift (Figures~\ref{fig:6.1}(b)) required setting the bandwidth and kernel function whereas completely unsupervised methods such as normalized cuts (Figures~\ref{fig:6.1}(c)) failed to converge. Clearly, the present method performs better as compared to the unsupervised methods tested. The overall performance of the present algorithm is {78\% and 75.6\%} on the Dice score for the BraTS 2013 and BraTS 2015 dataset, respectively. The 95\% Hausdorff distance for the two datasets are {13.61 and 14.05}, respectively. Based on the segmentation results, we also note that the present approach significantly minimizes oversegmentation as compared to most of the segmentation approaches reported in Figure~\ref{fig:6.1}. This is due to the TV regularization in the max-flow formulation that penalizes the oscillations in the pixel intensity due to random noise. We also compare the computational cost of the unsupervised algorithms that converged within a reasonable time and is summarized in Figure~\eqref{fig:7.1}. Although our method is slower than k-means and GMM, it performs reasonably fast as compared to mean-shift and spatially constrained GMM.   

The Dice score and the Hausdorff distance of all the methods including the supervised as well as unsupervised methods tested on the BraTS 2013 dataset are summarized in Figures~\ref{fig:8.1}(a)~\&~\ref{fig:8.1}(b). Interestingly, the present approach outperforms most of the supervised algorithms, both in terms of the average values of the Dice score and the Hausdorff distance (black squares inside the box plot indicate the average). The results suggest more than 90\% improvement in the Dice score and more than 56\% reduction in the Hausdorff distance when compared to the unsupervised methods. In terms of the supervised approaches, the present method performs statistically similar to the best performing method (Zhao(I) in \cite{menze2015}). For the BraTS 2015 dataset, no test results were available at the time of writing the manuscript, so we only compare with the training results of the algorithms as reported in the MICCAI 2015 proceedings \cite{menze2015multimodal}. These results are summarized in Table~\ref{table1}. Indeed, most of the deep learning based method outperform the results of the present algorithm. Nonetheless, the Dice scores for all the supervised methods reported in Table 1 were calculated on the training dataset where the training sizes were more than 85\% of the whole dataset \cite{wang2018deepigeos}.

\begin{table}
	\caption{Dice score comparison for the BRATS 2015 dataset}
	\begin{tabular}{lll}
		\hline
		Approach Summary                            & Dice score                  & Author  \\ \hline
		Unsupervised max-flow                       & 75.6$\pm$10.5  & Iquebal \\
		Generative with shape prior                & 77$\pm$19      & Agn  \cite{menze2015}   \\
		Generative-Discriminative  & 83$\pm$7.5   & Bakas \cite{menze2015}   \\
		Expectation Maximization                             & 68                          & Haeck \cite{menze2015}   \\
		Random Forests                             & 84                          & Maier \cite{menze2015}   \\
		Random Forests                             & 80.7                        & Malmi \cite{menze2015}   \\
		Conditional Random Fields                  & 82    & Meier \cite{menze2015}   \\
		Conv. Neural Networks               & 88                          & Havaei\cite{menze2015}  \\
		Conv. Neural Networks              & 81$\pm$15                          & Dvorak \cite{menze2015} \\
		Conv. Neural Networks              & 86                          & Pereira \cite{menze2015}\\
		Conv. Neural Networks                             & 67                          & Rao  \cite{menze2015}   \\
		Conv. Neural Networks                               & 81.41$\pm$9.6  & Vaidhya \cite{menze2015}\\
		Conv. Neural Networks                              & 87.55$\pm$6.72 & Wang \cite{wang2018deepigeos}   \\ \hline
	\end{tabular}\label{table1}
\end{table}


\subsection{Defect concentration in additively manufactured components}

Recent advances in the manufacturing technologies, especially metal additive manufacturing (i.e., layer-by-layer deposition of metal powder to fabricate complex free-form surfaces) have revolutionized the landscape of fabricating industrial components and parts. Despite the capability of AM to fabricate components with minimum time and material waste, the overall functional integrity of AM components is considered much inferior to those realized with conventional manufacturing process chains, especially under real-world dynamic loading conditions \cite{attar2014manufacture}. Defects, such as pores, undiffused metal powder, geometric distortions, surface cracks, and non-equilibrium microstructures significantly deteriorates the mechanical performance and overall functional integrity of the components. 

Concentration of defects in AM components is largely affected by the parameters of the AM process, e.g., laser power, laser scanning speed \cite{attar2014manufacture}. Controlling the process parameters can help manufacturers realize components with minimum defects and superior functional integrity \cite{tammas2017influence}. \textit{In situ} imaging technologies allow monitoring and detection of the defects induced in AM components. However, detection of defects using \textit{in situ} imaging technologies has its own limitations: (a) recording high-resolution images are costly, and therefore, only limited data is available, (b) uncertainties in the shape and morphology of defects, (c) low signal to noise ratio, etc. 

\begin{figure*}[!t]
	\centering{}
	\includegraphics[width=0.6\textwidth]{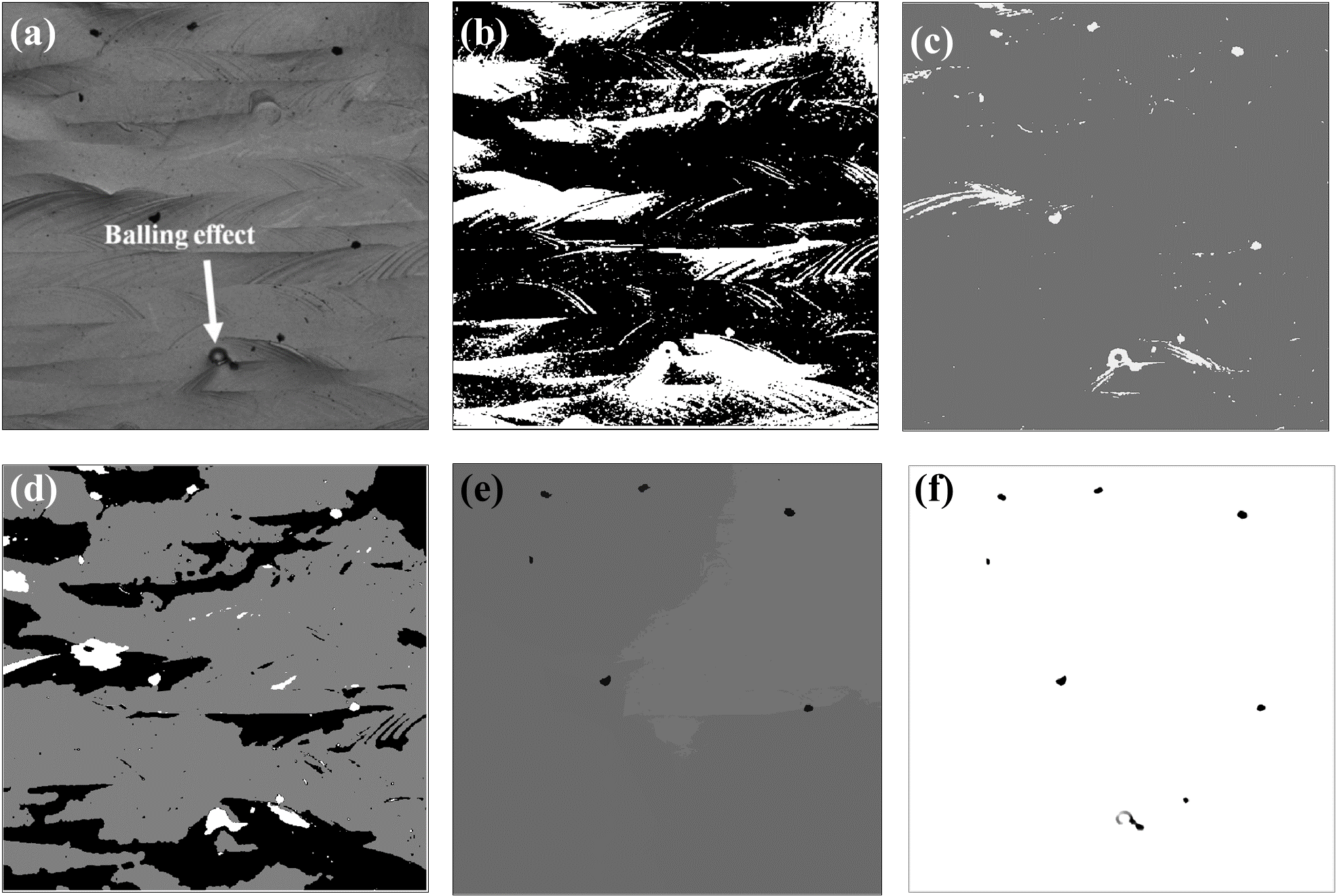}
	\caption{Comparative results of different algorithms tested for the segmentation of defects. (a) Original image for sample A (b) k-means with 2 clusters (c) Gaussian mixture model with expectation maximization (d) spatially constrained Gaussian mixture model with k-means initialization and (e) mean shift (f) the proposed method.}
	\label{fig:9.1}
\end{figure*}

\begin{figure*}[!t]
	\centering
	\includegraphics[width=0.6\textwidth]{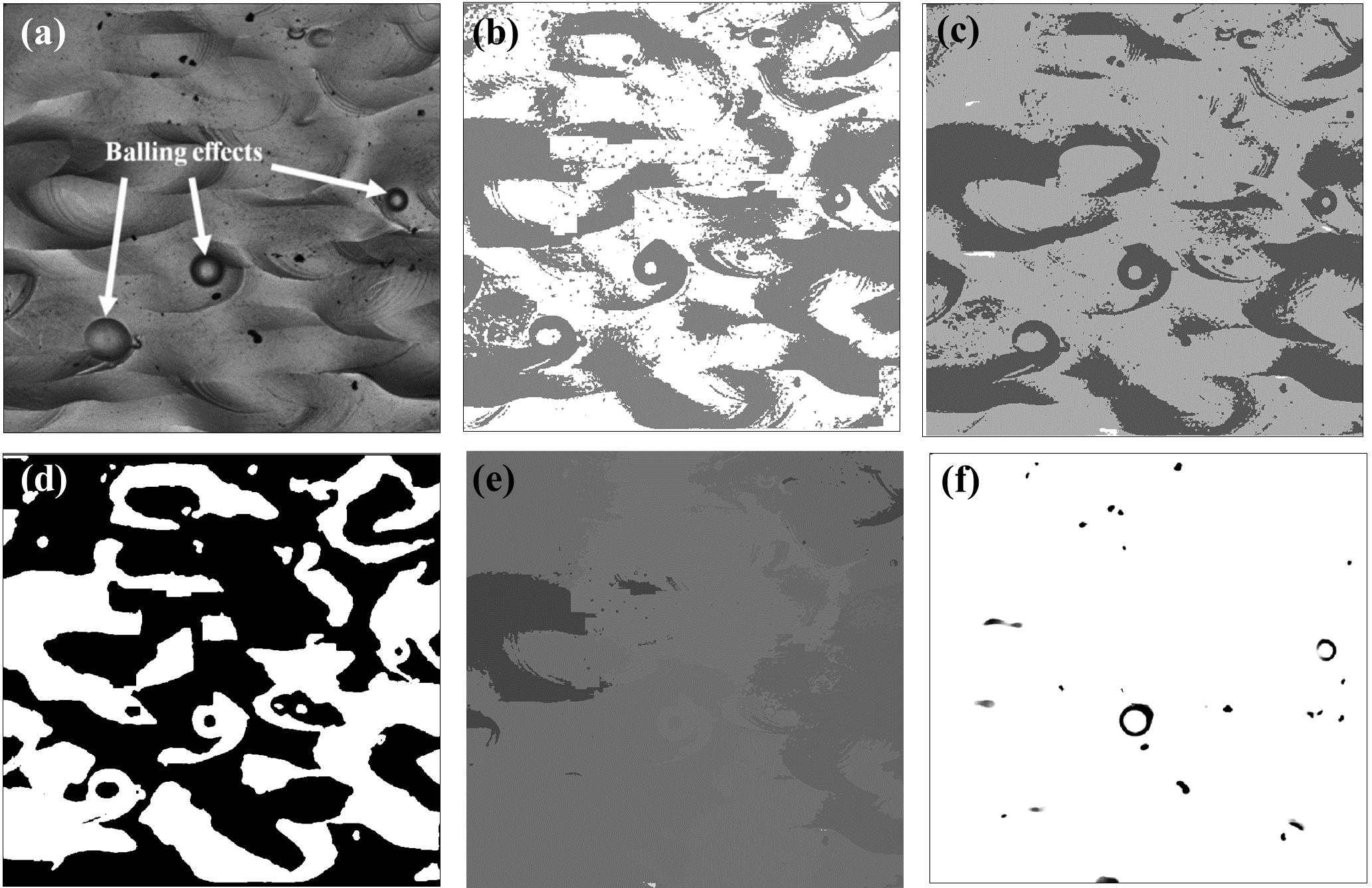}
	\caption{Comparative results of different algorithms tested for the segmentation of defects. (a) Original image for sample B (b) k-means with 2 clusters (c) Gaussian mixture model with expectation maximization (d) spatially constrained Gaussian mixture model with k-means initialization and (e) mean shift (f) the proposed method.}
	\label{fig:10.1}
\end{figure*}

In this case study, we employ the present approach to determine the concentration of defects on two different AM components that were fabricated with different process parameters using a laser-based AM process called selective laser melting (SLM). The process parameters investigated in this case study are laser power, laser scanning speed, and relative density of the AM components. For the first component (sample A, Figure~\ref{fig:9.1}(a)) the process parameters were set to: laser power = 165 W, laser scanning speed = 138 mm/s, and relative density = 99.5\%. For the second component (sample B, Figure~\ref{fig:10.1}(a)) these parameters were set to: laser power = 85 W, laser scanning speed = 71 mm/s, and relative density = 96.4\%. The images of the AM components were recored using an SEM (see \cite{attar2014manufacture} for details) and the defects were manually annotated to get the ground truth. 

We mainly focus on the concentration of two different types of defects namely, pores and balling effect. Pores are essentially voids on the surface that are formed when gas particles trapped in the melt-pool (liquefied metal during deposition) escape. These are generally tracked by observing the dark spherical/oval shaped features on the surface. In contrast, balling effect is a complex metallurgical process that originates due to sub-optimal process parameters during the SLM process as well as the properties of the material powder such as the relative density of AM components. It causes the liquid scan track during SLM to break and result in the formation of spherical particles that eventually get trapped, causing inhomogeneous deposition of the powder in the next build layer. Authors in \cite{attar2014manufacture} showed that by increasing the laser power, the concentration of defects due to the balling effect decreases. Controlling pores as well as the balling effect is critical to avoid significant costs, as these defects may reduce the fatigue strength of material by more than 4 times, resulting in early, unexpected failure \cite{iquebal2017longitudinal}. 

Figure~\ref{fig:9.1}(a) and \ref{fig:10.1}(a) shows the representative surface from sample A and sample B, respectively. We note that sample A with high laser power (165 W) has relatively lower defect concentration as compared to sample B with low laser power (85 W). As the data size is limited, we compare our segmentation results only with that of the unsupervised segmentation methods. To keep the comparison fair, we ignore the methods that either required user inputs to define the foreground and background or did not converge in a reasonable time. The segmentation results for the first sample are presented in Figures~\ref{fig:9.1}(b-f). The effect of the noise (resulting from the shadow and underexposure) is clearly reflected in the results obtained from k-means clustering approach as shown in Figure~\ref{fig:9.1}(b). Although, there is some improvement in the segmentation obtained from GMM and SC-GMM as shown in Figure~\ref{fig:9.1}(c-d), yet the effect of noise is evident from the resulting oversegmentation. The segmentation from the mean shift algorithm (Figure~\ref{fig:9.1}(e)) is able to detect most of the pores (with 58.6\% Dice score) but fails to detect the balling effect. Finally, the segmentation obtained with the proposed method is able to identify all the regions containing pores as well as the balling effect with a Dice score of 70.56\% and Hausdorff distance of 32---approximately 86\% smaller than the rest of the unsupervised segmentation methods (Figure~\ref{fig:9.1}(f)).

\begin{table}[]
	\caption{Average Dice score and Hausdorff measure of various unsupervised approaches implemented for defect segmentation.}
	\begin{tabular}{ccc}
		\hline
		Approach                  & Dice score & Hausdorff measure\\ \hline
		Unsupervised max-flow     &     70.56  &                 32  \\
		Mean shift                & 	58.6   & 245           \\
		GMM    &     22.16       &    287               \\
		SC-GMM &     2.6       &      219             \\ 
		k-means                   & 1.4           & 321 \\ \hline
	\end{tabular}\label{table2}
\end{table}

In the second sample, noise is significantly higher as compared to sample A, mainly due to the rougher surface morphology. Three instances of balling effect were identified manually as shown by the arrows in Figure~\ref{fig:10.1}(a), alongside multiple pores. Clearly, the segmentation results obtained from k-means (Figure~\ref{fig:10.1}(b)), GMM (Figure~\ref{fig:10.1}(c)), SC-GMM (Figure~\ref{fig:10.1}(d)) and mean shift (Figure~\ref{fig:10.1}(e)) mostly capture the noise present in the original image. In contrast, the proposed method is able to selectively segment the defects while significantly reducing the oversegmentation (Figure~\ref{fig:10.1}(f)). However, the algorithm is able to identify only two out of the three areas showing balling effect. Nonetheless, the segmentation is a significant improvement over the standard state-of-the-art unsupervised segmentation methods. A summary of the Dice score and the Hausdorff measure of all the unsupervised approaches is presented in Table~\ref{table2}.

As mentioned earlier, estimating defect concentration in AM can help understand the effect of various process parameters on the build quality, and therefore determine the optimal parameter settings \cite{iquebal2017longitudinal,attar2014manufacture}. Via visual inspection, we were able to verify the experimental observations \cite{attar2014manufacture} that the increase in the laser power results in the decrease in defect concentration. By estimating the area fraction of defects from segmented images, we note that the defect concentration on sample A is $\sim$1\% and sample B is $\sim$ 4.39\%. This is in accordance with the experimental observations. From a process standpoint, the present approach may be integrated with an experimental design strategy to determine the optimal process parameter settings that would minimize the defect concentration.   

\section{Conclusions}

Advances in microscopy and functional imaging technologies open exciting opportunities for fast and on the fly detection/segmentation of ROIs using image snapshots and streams. However, the uncertainty associated with the shape and location of ROIs renders the task of generating annotations and atlases extremely costly and time-consuming. Although considerable research exists in the image segmentation literature, a majority of the methods rely on huge training datasets or require manual intervention to set the parameters \textit{apriori}. In contrast, only a handful of unsupervised approaches have been reported in the literature, many of which are computationally complex (e.g., normalized cuts) or require partial supervision (e.g., continuous max-flow). 

In the present work, we developed an approach to consistently estimate the flow capacity parameters leading to a fully unsupervised image segmentation approach. Our framework is based on iteratively estimating the image labels using a continuous max-flow approach followed by the MAP estimation of the flow capacities by considering an MRF prior over the flow field. In the sequel, we presented results to validate the consistency of the posterior distribution of the flow capacities to ensure that the estimated flow capacities are consistent under the MRF prior proposed in this work. Segmentation results on two distinct real-world case studies, including brain tumor segmentation (BraTS) using FLAIR-MR scans and defect segmentation in additively manufactured components using SEM images are presented. An extensive comparison with the state-of-the-art unsupervised as well as supervised algorithms suggests that the present method outperforms all the unsupervised as well as most of the supervised algorithms tested. More specifically, we note that the present approach results in more than 90\% improvement in the Dice score and more than 56\% reduction in the Hausdorff distance when compared to the unsupervised methods. Future works are focused on the segmentation of 3D images as well as a more comprehensive Bayesian model to estimate the parameters used in the MRF prior. 

\ifCLASSOPTIONcompsoc
  \section*{Acknowledgments}
\else
  \section*{Acknowledgment}
\fi

The authors would like to acknowledge the kind support from the National Science Foundation through award ECCS 1547075 and the X-Grants, part of the Texas A\&M President's Excellence Fund at Texas A\&M University.

\ifCLASSOPTIONcaptionsoff
  \newpage
\fi



%

\bibliographystyle{IEEEtran}
\bibliography{references}

%

%
%
%




\vskip -4em
\begin{IEEEbiography}[{\includegraphics[width=1in,height=1.25in, keepaspectratio]{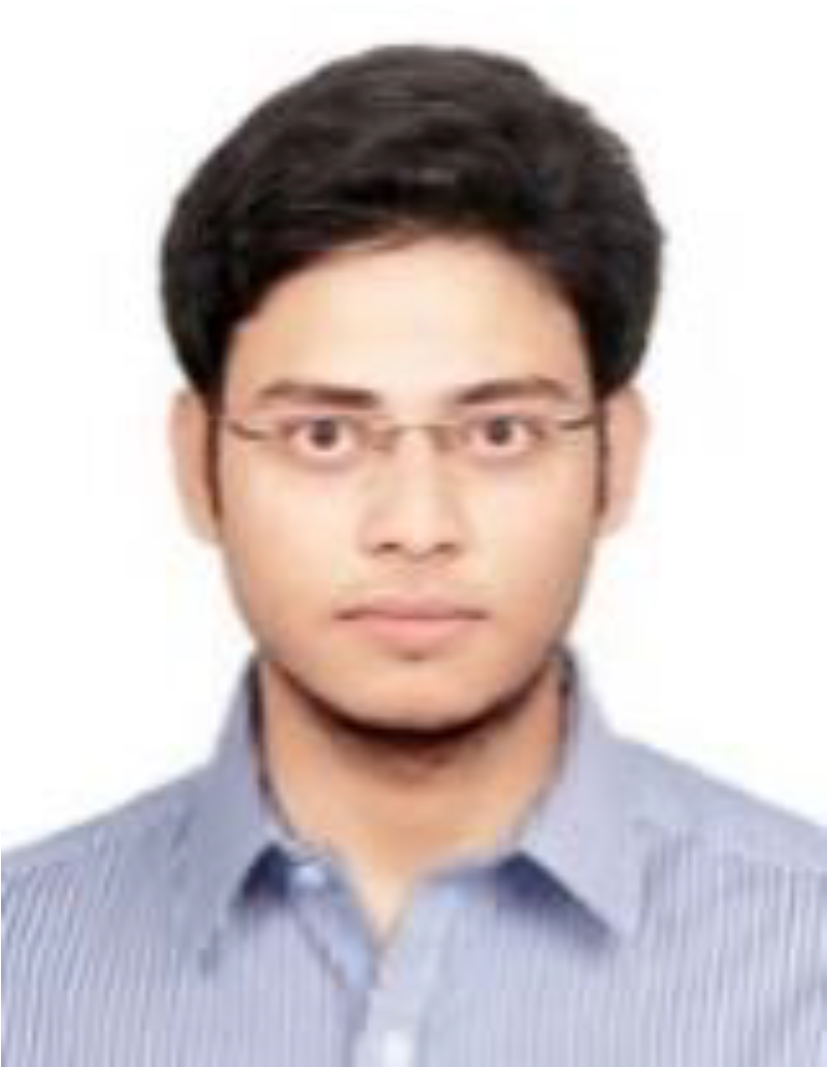}}]{Ashif S. Iquebal }
	is a Ph.D. candidate in the department of Industrial and Systems Engineering at Texas A\&M University, College Station, TX, USA. He has obtained his Master's degree in Statistics from Texas A\&M University in 2019 and Bachelor's in Industrial and Systems Engineering with minor in Mathematics and Computing from Indian Institute of Technology, Kharagpur, India in 2014. His research interests include streaming data analysis, image-based unsupervised anomaly detection in health-care and manufacturing systems, and active learning for optimal experimental design. He is a student member of IEEE, Institute of Industrial and Systems Engineers (IISE), and the Institute for Operations Research and the Management Sciences.

\end{IEEEbiography} 
	\vskip -5em	
\begin{IEEEbiography}[{\includegraphics[width=1in,height=1.25in, keepaspectratio]{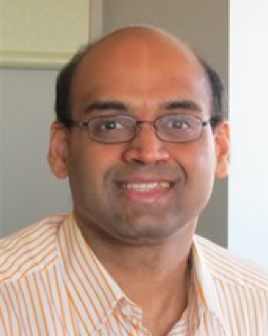}}]{Satish Bukkapatnam}
serves as Rockwell International Professor with Department of Industrial and Systems Engineering department at Texas A\&M University and the Director of the Texas A\&M Engineering Experimentation Station, Institute for Manufacturing Systems and an affiliate faculty appointment at The \'{E}cole Nationale Sup\'{e}rieure Arts et M\'{e}tiers, France. His research addresses the harnessing of high-resolution nonlinear dynamic information, especially from wireless MEMS sensors, to improve the monitoring and prognostics, mainly of ultraprecision and nanomanufacturing processes and machines, and cardiorespiratory processes. His research has led to 151 peer-reviewed publications, five pending patents, 14 completed Ph.D. dissertations, \$5 million in grants as PI/Co-PI from the National Science Foundation, the U.S. Department of Defense, and the private sector, and 17 best-paper/poster recognitions. He is a fellow of the IISE and the Society of Manufacturing Engineers, and he has been recognized with Oklahoma State University Regents distinguished research, Halliburton outstanding college of engineering faculty, IISE Boeing technical innovation, IISE Eldin outstanding young industrial engineer, and SME Dougherty outstanding young manufacturing engineer awards. He currently serves as the editor of the IISE Transactions, Design and Manufacturing Focused Issue. He received his master's and Ph.D. degrees from the Pennsylvania State University and undergraduate degree from S.V. University, Tirupati, India.
\end{IEEEbiography} 

\end{document}